\documentclass[11pt, a4paper,logo,  copyright, nonumbering]{paddleocr}
\usepackage[numbers]{natbib}
\usepackage{dblfloatfix}
\usepackage{ulem}
\usepackage{float}
\usepackage{tocloft}
\usepackage{caption}
\usepackage{dramatist}
\usepackage{xspace}
\usepackage{array}
\usepackage{pifont} %
\usepackage{multirow}
\usepackage{geometry}
\usepackage{makecell}
\usepackage{tcolorbox}
\usepackage{xltabular}
\usepackage{titlesec} 
\usepackage{longtable}
\usepackage[hang, flushmargin]{footmisc}
\usepackage{booktabs} 
\usepackage{xcolor}   
\usepackage[table]{xcolor}
\definecolor{lightpink}{RGB}{255, 182, 193}
\definecolor{iccvblue}{rgb}{0.21,0.49,0.74}
\usepackage{wrapfig}
\usepackage{amssymb}
\usepackage[ruled,vlined,linesnumbered,algo2e]{algorithm2e} 

\hypersetup{
    colorlinks=true,
    linkcolor=blue,
    citecolor=blue,
    filecolor=blue,
    urlcolor=blue,
    pdfpagemode=UseNone,
    pdfstartview=FitH
}
\usepackage{threeparttable} 
\usepackage{soul}
\usepackage{amsfonts}
\usepackage{amsmath}
\usepackage{amssymb}
\usepackage{lineno}
\usepackage{multirow}
\usepackage{adjustbox}

\usepackage{CJKutf8}
\usepackage{subcaption}
\usepackage{setspace}

\usepackage{dsfont}
\usepackage{array} %
\usepackage{tabularx} %
\usepackage{xcolor} %
\usepackage{tabularx}
\usepackage{booktabs}

\usepackage{lipsum}  %
\usepackage{multicol} %
\usepackage{makecell} 
\usepackage{listings}
\usepackage{xcolor}

\usepackage{tablefootnote}
\usepackage{hyperref}

\lstdefinelanguage{json}{
    basicstyle=\ttfamily\footnotesize,
    numbers=left,
    numberstyle=\tiny\color{gray},
    stepnumber=1,
    numbersep=8pt,
    showstringspaces=false,
    breaklines=true,
    frame=lines,
    backgroundcolor=\color{gray!10},
    morestring=[b]",
    literate=
     *{0}{{{\color{black}0}}}{1}
      {1}{{{\color{black}1}}}{1}
      {2}{{{\color{black}2}}}{1}
      {3}{{{\color{black}3}}}{1}
      {4}{{{\color{black}4}}}{1}
      {5}{{{\color{black}5}}}{1}
      {6}{{{\color{black}6}}}{1}
      {7}{{{\color{black}7}}}{1}
      {8}{{{\color{black}8}}}{1}
      {9}{{{\color{black}9}}}{1}
}

\makeatletter
\def\@BTrule[#1]{%
  \ifx\longtable\undefined
    \let\@BTswitch\@BTnormal
  \else\ifx\hline\LT@hline
    \nobreak
    \let\@BTswitch\@BLTrule
  \else
     \let\@BTswitch\@BTnormal
  \fi\fi
  \global\@thisrulewidth=#1\relax
  \ifnum\@thisruleclass=\tw@\vskip\@aboverulesep\else
  \ifnum\@lastruleclass=\z@\vskip\@aboverulesep\else
  \ifnum\@lastruleclass=\@ne\vskip\doublerulesep\fi\fi\fi
  \@BTswitch}
\makeatother

\addto\extrasenglish{
}

 {\begin{list}{}%
         {\setlength{\leftmargin}{#1}}%
         \item[]%
 }
 {\end{list}}

\reportnumber{001} %

\renewcommand{\today}{}

\newcommand{\printfnsymbol}[1]{%
  \textsuperscript{\@fnsymbol{#1}}%
}

\title{\centering HPD-Parsing: Hierarchical Parallel Document Parsing}

\author[*]{
\small
Shu Wei$^*$, Jingjing Wu$^*$, Lingshu Zhang$^*$, Qunyi Xie$^{\dagger}$, Hao Zou, Le Xiang, Xu Fan, Yangliu Xu, 
\vspace{-0.4cm}
\\
\small
Manhui Lin, Xiaolong Ma, Cheng Cui, Tengyu Du, YY
\vspace{0.1cm}
\\
\small
$^*$Equal contribution,
$^{\dagger}$Project Leader
\\
\vspace{0.1cm}
{\small
  \raggedright{  

  \small
  \hspace{11em}  
  \includegraphics[height=1.0em]{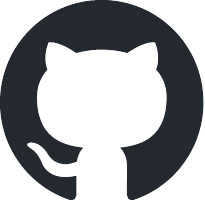}  
  \quad \url{https://github.com/PaddlePaddle/PaddleOCR} \\
  \hspace{1.5em}  
  \small
  \includegraphics[height=1.0em]{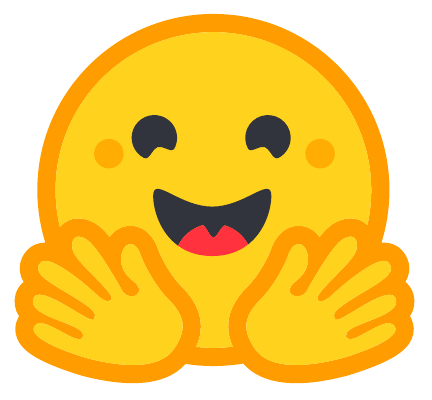} \quad \url{https://huggingface.co/PaddlePaddle/HPD-Parsing} \\
  \small
  \hspace{-0em}  
  \includegraphics[height=0.8em]{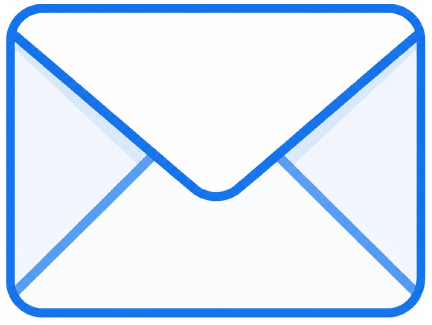}
  {\quad paddleocr@baidu.com}
  }
}
}

\renewcommand{\phi}{\varphi}

\renewcommand{\leq}{\leqslant}
\renewcommand{\geq}{\geqslant}

\renewcommand{\epsilon}{\varepsilon}
\renewcommand{\imath}{\mathrm{i}}

\newlength{\restsubwidth}
\newlength{\restsubheight}
\newlength{\restsubmoreheight}
\newcommand{\rest}[2]{%
        \settowidth{\restsubwidth}{\ensuremath{#2}}
        \settoheight{\restsubheight}{\ensuremath{{}_{#2}}}
        \ensuremath{{#1\hskip 0.5pt}_{\vrule\kern2pt\parbox[b][%
        4pt][b]{\the\restsubwidth}{%
                        \ensuremath{{}_{#2}}}}}
        }

\vspace{-10cm} 
\begin{abstract}
Efficient teamwork typically combines global coordination with parallel execution, a principle not yet fully reflected in unified Vision-Language Model (VLM)-based document parsers. Existing unified parsers process an entire page jointly but generate its output through a single token-by-token autoregressive trajectory, creating a sequential bottleneck that grows with document length. Such full-page sequential generation overlooks a key property of document parsing: layout must be analyzed globally, whereas block content can be parsed in parallel. Based on this observation, we introduce \textbf{HPD-Parsing}, which replaces full-page autoregressive generation with a \textbf{H}ierarchical \textbf{P}arallel \textbf{D}ecoding paradigm. A main layout branch organizes the overall document structure and dynamically assigns block-level content decoding to concurrent branches, while progressive multi-token prediction (P-MTP) further reduces the decoding steps within each branch. Experiments on public benchmarks show that HPD-Parsing achieves 4,752 tokens per second, delivering \textbf{$2.62\times$} the throughput of the fastest existing document parsing model and \textbf{$3.06\times$} that of the vanilla autoregressive baseline, while maintaining competitive parsing accuracy. These results establish hierarchical parallel decoding as an effective alternative to full-page autoregressive generation, opening a new direction for efficient unified document parsing.
\end{abstract}

\vspace{-0.5cm} 

\begin{document}

\maketitle
\vspace{-0.3cm} 
\begin{figure}[h]
    \centering
\includegraphics[width=1\linewidth]{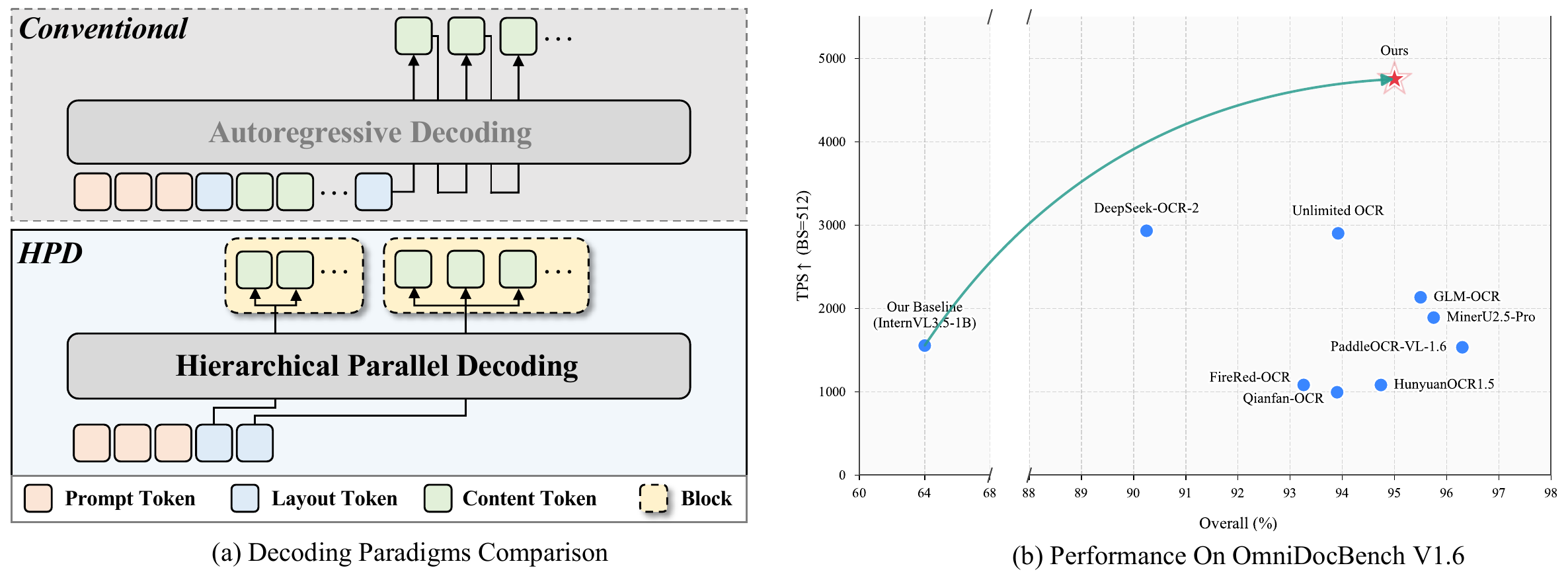}
    \vspace{-0.7cm}
    \caption{Comparison of decoding paradigms and inference efficiency. (a)  Conventional autoregressive decoding generates the full-page output sequentially, whereas HPD-Parsing combines layout-coordinated parallel decoding across document blocks with P-MTP. (b) HPD-Parsing delivers a 3.06$\times$ throughput improvement over the autoregressive baseline, while achieving state-of-the-art inference efficiency with competitive parsing accuracy.
}
    \label{fig:teaser}
\end{figure}
\newpage

\section{Introduction}

Document parsing serves as a fundamental component in a wide range of downstream applications, including information extraction, document retrieval, and Retrieval-Augmented Generation (RAG), placing high demands on parsing accuracy. Compared with traditional pipeline-based systems, recent Vision-Language Model (VLM)-based parsers have substantially improved accuracy and generality by jointly modeling diverse document elements within a unified framework. However, document pages often contain dense content and require long structured outputs, causing VLM-based parsers to incur substantially higher decoding costs than traditional systems because of token-by-token autoregressive generation. As document workloads continue to grow, improving inference efficiency while preserving parsing quality has therefore become an increasingly important problem.

Existing VLM-based document parsers can be broadly categorized into pipeline-based and unified methods. Pipeline-based methods~\cite{cui2025paddleocr,cui2026paddleocr,wang2024mineru,wang2026mineru2,li2025monkeyocr,yin2026youtu,feng2025dolphin,feng2026dolphin,duan2026glm} typically combine specialized components for layout analysis and VLM-based content decoding, enabling flexible region-wise processing with inherently fragmented workflows. Unified methods~\cite{wu2026firered,dong2026qianfan,team2025hunyuanocr,li2025dots,duan2026glm,chen2025ocean,blecher2024nougat,taghadouini2026lightonocr,wei2025deepseek,wei2026deepseek,chen2025logics}, by contrast, integrate layout parsing and content recognition into a unified sequence generation framework, enabling joint optimization within a single model.  Nevertheless, VLM-based decoding in existing parsers is still predominantly autoregressive, requiring document content to be generated token by token. This serial generation process leads to high inference latency and limited throughput, making it difficult to satisfy the efficiency requirements of large-scale document parsing. 

To alleviate the inference bottleneck, recent studies have pursued two complementary directions. One line of work reduces attention cost by shortening the effective context. DeepSeek-OCR~\cite{wei2025deepseek,wei2026deepseek} compresses visual tokens to reduce redundancy in high-resolution document inputs, while Unlimited OCR~\cite{yin2026unlimited} introduces Reference Sliding Window Attention (R-SWA) to bound the textual KV cache during long-sequence decoding. A second line of work shortens the generation path by increasing decoding parallelism. Youtu-Parsing~\cite{yin2026youtu} introduces query- and token-level parallelism for region-wise recognition, GLM-OCR~\cite{duan2026glm} employs Multi-Token Prediction (MTP) to advance multiple tokens per decoding step, and HunyuanOCR-1.5~\cite{li2026hunyuanocr} adapts DFlash for parallel draft generation. These advances offer complementary efficiency improvements, while the broader decoding paradigm for full-page document generation remains underexplored.

Beyond existing acceleration strategies, we observe a fundamental mismatch in unified document parsing: while page-level parsing requires global coordination, content decoding is largely localized. The global layout determines the spatial structure, region relationships, and reading order of a document, whereas the detailed content of each region is primarily grounded in its corresponding visual evidence and exhibits limited dependence on distant regions. This makes a single full-page autoregressive trajectory unnecessarily sequential. Moreover, the strong visual grounding of localized content yields relatively predictable generation trajectories, making document parsing well suited to multi-token prediction. Motivated by this insight, we propose HPD-Parsing, a high-throughput document parser built on \emph{Hierarchical Parallel Decoding}. A main layout branch performs global coordination and dynamically decomposes the conventional single decoding trajectory into concurrent local content branches with shared-prefix KV cache reuse. Within each branch, P-MTP~\cite{xiang2026p} further reduces decoding steps by predicting multiple future tokens at each iteration. Collectively, these designs shorten the effective sequential decoding path across branches and within each branch while preserving a unified generation framework. To maintain parsing accuracy during this paradigm transition, we further adopt staged adaptation and automated difficulty-aware data curation. As shown in Figure~\ref{fig:teaser}(b), HPD-Parsing achieves best throughout while maintaining competitive parsing accuracy.

In summary, the key contributions of HPD-Parsing are as follows:
\begin{itemize}

\item \textbf{Hierarchical Parallel Decoding for High-Throughput Document Parsing.}
We introduce Hierarchical Parallel Decoding (HPD), a new decoding paradigm that restructures full-page autoregressive generation into globally coordinated, localized parallel decoding. A main layout branch performs global coordination and dynamically decomposes the conventional single decoding trajectory into concurrent content branches, each responsible for a localized document region. Within each branch, P-MTP further reduces the number of decoding steps by predicting multiple future tokens at each iteration. Together with shared-prefix KV cache reuse, HPD substantially shortens the effective sequential decoding path along both branch and token dimensions.

\item \textbf{Staged Adaptation with Automated Difficulty-Aware Data Curation.}
We develop a staged adaptation strategy that transfers conventional autoregressive document parsing capabilities to the proposed hierarchical parallel decoding paradigm while preserving parsing accuracy. The strategy is supported by an automated difficulty-aware data curation pipeline that integrates large-scale data collection, model-assisted annotation, difficulty estimation, and balanced sampling. By progressively adapting the model and emphasizing challenging samples, the training framework mitigates the accuracy degradation caused by the transition to parallel decoding with minimal manual annotation effort.

\item \textbf{State-of-the-Art Throughput with Competitive Parsing Accuracy.}
HPD-Parsing achieves state-of-the-art inference efficiency on OmniDocBench v1.6, reaching a peak throughput of \textbf{4,752 Tokens Per Second (TPS)}. It delivers $1.62\times$ the throughput of the fastest existing document parsing model and more than $3.06\times$ that of its autoregressive baseline, while maintaining competitive parsing accuracy. These results demonstrate that document parsing can be effectively executed through global layout coordination and localized parallel decoding rather than a single sequential generation trajectory.
\end{itemize}

\section{Related Work}

\subsection{VLM-based Document Parsing}

Existing VLM-based document parsers can be broadly categorized into pipeline-based and unified methods according to their system organization. Pipeline-based methods~\cite{cui2025paddleocr,cui2026paddleocr,wang2024mineru,wang2026mineru2,li2025monkeyocr,yin2026youtu,feng2025dolphin,feng2026dolphin,duan2026glm} decompose document parsing into a sequence of specialized stages, typically including page-level layout analysis followed by VLM-based recognition or content decoding for localized regions. This modular design enables flexible optimization for different document elements and has demonstrated strong parsing performance in systems such as the PaddleOCR-VL and MinerU series. However, coordinating multiple components also increases system complexity and may propagate localization errors to subsequent recognition stages.

Unified methods~\cite{wu2026firered,dong2026qianfan,team2025hunyuanocr,li2025dots,duan2026glm,chen2025ocean,blecher2024nougat,taghadouini2026lightonocr,wei2025deepseek,wei2026deepseek,chen2025logics,yin2026unlimited,li2026hunyuanocr} instead formulate layout parsing, reading-order prediction, and content recognition within a unified sequence generation framework. Recent studies have improved this paradigm through large-scale data curation~\cite{lyu2024structextv3,li2025dots,wu2026firered,dong2026qianfan}, multi-task supervision and output formatting~\cite{team2025hunyuanocr}, and reinforcement learning~\cite{taghadouini2026lightonocr,wu2026firered,chen2025logics}. Despite substantial progress in parsing accuracy, VLM-based decoders in both paradigms remain predominantly autoregressive. Whether applied to localized regions or an entire page, document content is generally generated token by token, resulting in substantial inference latency for text-dense documents with long structured outputs.

\subsection{Acceleration Strategies for Document Parsing}

Recent studies have accelerated VLM-based document parsing along two complementary dimensions: reducing per-step computation and shortening the sequential decoding path. DeepSeek-OCR~\cite{wei2025deepseek,wei2026deepseek} compresses visual tokens to reduce the attention cost of high-resolution document inputs, while Unlimited OCR~\cite{yin2026unlimited} introduces Reference Sliding Window Attention (R-SWA) to maintain a bounded textual KV cache during long-sequence generation. Both approaches improve efficiency by limiting the effective context processed at each decoding step. Complementary efforts increase parallelism within the decoding process to reduce the number of sequential decoding rounds. Youtu-Parsing~\cite{yin2026youtu} introduces query- and token-level parallelism for region-wise content recognition, while GLM-OCR~\cite{duan2026glm} employs Multi-Token Prediction (MTP) to advance multiple tokens per decoding step. HunyuanOCR-1.5~\cite{li2026hunyuanocr} further adapts DFlash to OCR decoding, enabling parallel draft generation for faster autoregressive inference. Collectively, these methods reduce the cost of context processing or accelerate token generation within the conventional autoregressive pipeline. Yet, how to restructure full-page document generation beyond the conventional autoregressive trajectory remains largely unexplored.

\section{Methodology}
\subsection{From Sequential Bottleneck to Hierarchical Parallel Decoding}



\begin{wrapfigure}[20]{r}{0.5\linewidth}
    \centering
    
    \includegraphics[width=\linewidth]{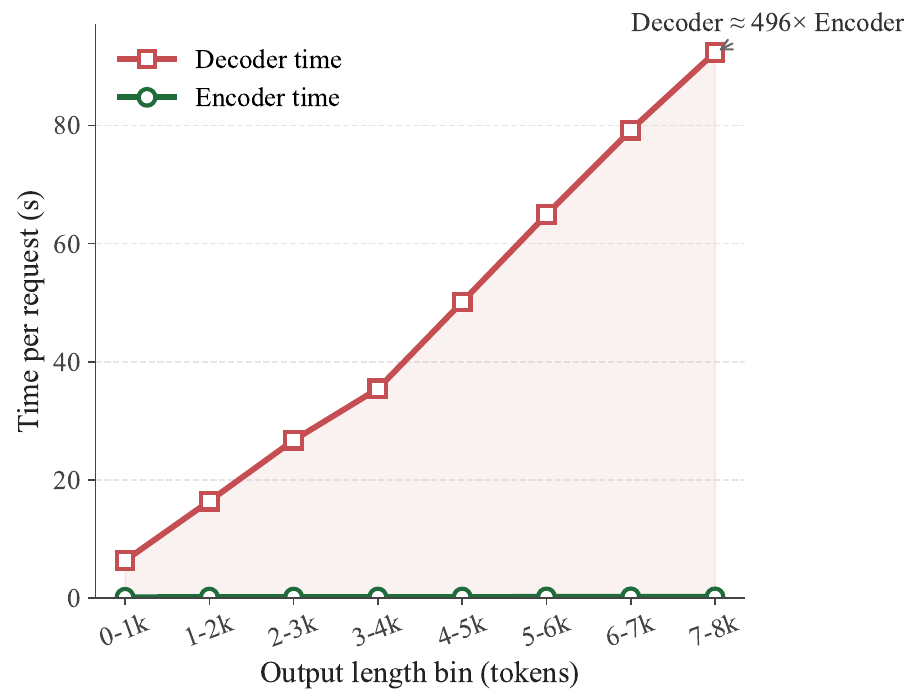}
    \caption{Profiling results on a standard InternVL3.5-1B-based autoregressive parsing model show that decoder latency increases rapidly with output length and becomes the dominant inference cost.}
    \label{fig:example}
\label{fig:motivation}
\end{wrapfigure}

Before introducing HPD-Parsing, we first analyze the inference bottleneck of standard autoregressive document parsing. For this purpose, we train a baseline document parser using InternVL3.5-1B as the backbone. The model follows the conventional unified VLM-based parsing pipeline: a document image is first encoded into visual tokens, after which the complete parsing result is generated token by token by the LLM decoder. As shown in Figure~\ref{fig:motivation}, we profile the encoder and decoder latency across samples grouped by output length using vLLM with an inference batch size of 16.

As the output length increases, encoder latency remains relatively stable despite moderate variation in the number of visual tokens. Decoder latency, by contrast, grows rapidly with sequence length and progressively dominates the overall inference cost. For samples with long outputs, decoding can take nearly 500 times longer than visual encoding. These results indicate that the primary efficiency bottleneck of unified document parsing lies not in full-page visual processing, but in the long sequential execution path imposed by token-by-token autoregressive decoding.

Overcoming this bottleneck requires reconsidering the organization of document generation. Although document parsing relies on global coordination to determine spatial structure, region relationships, and reading order, the detailed content of each region is largely localized. Text paragraphs, tables, and formulas are primarily grounded in their corresponding visual regions and typically exhibit limited dependence on the detailed content of distant regions. Therefore, global document structure must be coordinated jointly, but regional contents need not be generated sequentially along a single autoregressive trajectory.

This observation motivates a hierarchical decoding organization. A main layout branch establishes the global layout structure and reading order, while dynamically decomposing the conventional single decoding trajectory into concurrent local content branches. Each content branch decodes a localized region, with shared-prefix KV cache reuse avoiding redundant computation. Within each branch, strong grounding in local visual evidence further makes the generation trajectory more predictable, creating favorable conditions for Progressive Multi-Token Prediction (P-MTP) to reduce the number of autoregressive decoding steps. Based on these insights, we formulate document parsing as \emph{Hierarchical Parallel Decoding}: global layout coordination enables localized parallel decoding across content branches, while P-MTP further accelerates generation within each branch.

\subsection{Overview}

\begin{figure}[t]
    \centering
    \includegraphics[width=1\linewidth]{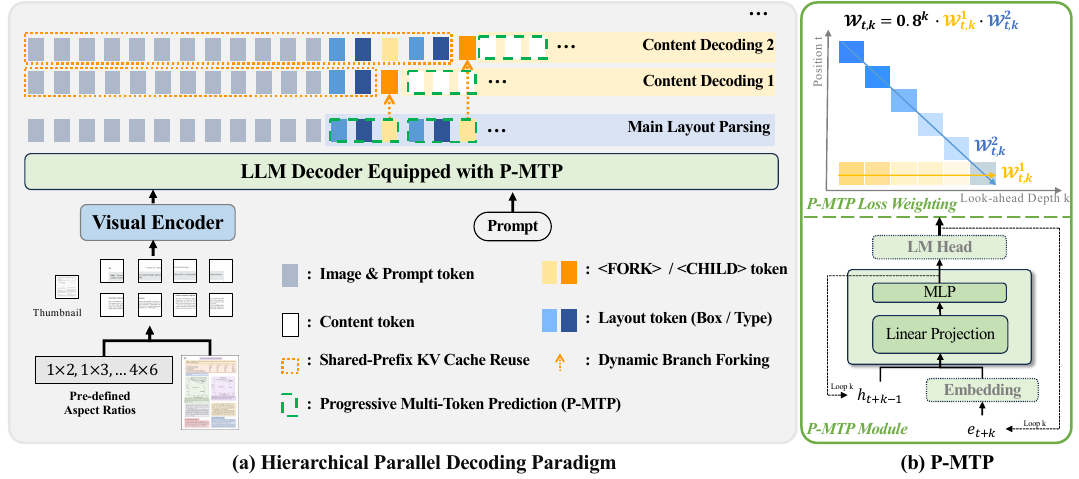}
    \caption{HPD-Parsing accelerates document parsing through a hierarchical parallel decoding paradigm that combines layout-coordinated branch decoding with P-MTP. In (a), the input document image is encoded into visual tokens under predefined aspect ratios. A main layout branch coordinates the global document structure and dynamically forks multiple content branches for concurrent localized decoding, while reusing the shared-prefix KV cache. Within each branch, P-MTP further shortens the decoding path through progressive look-ahead prediction. In (b), P-MTP comprises a lightweight  prediction module and a progressive loss-weighting strategy for supervising predictions at increasing look-ahead depths. 
    }
    \label{fig:framework}
\end{figure}

As illustrated in Figure~\ref{fig:framework}, HPD-Parsing adopts InternVL3.5-1B~\cite{wang2025internvl3} as its backbone, comprising a 0.3B InternViT visual encoder and a 0.8B LLM decoder. The visual encoder extracts high-resolution document representations, which are projected into the language space and processed by the decoder for unified document parsing. 
\noindent\textbf{Visual Encoder.} InternViT adopts dynamic tile-based cropping to efficiently preserve high-resolution details. During both training and inference, each input image is adaptively partitioned into up to 24 tiles according to its resolution and aspect ratio. Each tile is resized to $448 \times 448$ and independently encoded, after which the resulting visual tokens are fed into the LLM.

\noindent\textbf{LLM Decoder.} LLM Decoder is adapted from the Qwen3-0.6B architecture, constituting approximately 0.8B parameters of the overall framework. It consists of 28 Transformer layers with a hidden dimension of 1024 and an intermediate FFN size of 3072. To manage attention complexity, the decoder utilizes Grouped-Query Attention (GQA) configured with 16 query heads and 8 key-value heads. SwiGLU is adopted as the activation function coupled with RMSNorm. Built upon this decoder, HPD-Parsing replaces the conventional single autoregressive trajectory with layout-coordinated dynamic branch forking, enabling the main layout branch and localized content branches to decode concurrently. Our previously proposed P-MTP is further integrated into each branch to reduce the remaining autoregressive decoding steps. The layout-coordinated branch decoding mechanism and its integration with P-MTP are described in Section~\ref{sec:block} and Section~\ref{sec:p-mtp}, respectively.


\subsection{Layout-Coordinated Parallel Decoding}
\label{sec:block}

HPD-Parsing introduces layout-coordinated parallel decoding to exploit the structural locality of document pages and shorten the sequential execution path of full-page generation. Rather than generating the complete parsing result along a single autoregressive trajectory, it assigns different decoding roles to global layout coordination and localized content generation. A main layout branch sequentially establishes the document structure in reading order, while content branches are dynamically forked to decode individual layout regions concurrently.

\noindent\textbf{Decoding Roles and Sequence Formats.}
The hierarchical decoding process comprises two types of branches:
\begin{itemize}
\item \textbf{Layout branch.} The main layout branch generates a structured sequence following the natural reading order of the page. Each layout unit is represented by its category, normalized spatial coordinates, and a routing token \texttt{<FORK>}, which indicates that the corresponding regional content can be decoded by a separate branch.
\item \textbf{Content branch.} Each local content branch generates the content associated with one layout unit. Its input consists of the shared visual context and the structural prefix corresponding to that unit, followed by a transition token \texttt{<CHILD>} that marks the beginning of localized content generation.
\end{itemize}

\noindent\textbf{Branch-specific Sequence Supervision.} 
Based on the two sequence formats, each original document annotation is transformed into a layout-branch sequence and a set of block-specific content-branch sequences. For layout parsing, all valid tokens in the structural sequence are supervised. For content decoding, the visual context and block-specific structural prefix serve only as conditioning information, while supervision is restricted to the localized transcription following \texttt{<CHILD>}. The corresponding validity mask is defined as
\begin{equation} 
\label{eq:dec_mask}
    \mathcal{M}_t^{\text{dec}} = \mathbb{I}(t > t_{\langle\text{CHILD}\rangle}) ,
\end{equation}
where $t_{\langle\mathrm{CHILD}\rangle}$ denotes the position of \texttt{<CHILD>} in the content-decoding sequence. This masking strategy allows the structural prefix to condition localized generation without requiring the content branch to reproduce the shared context.

\noindent\textbf{Dynamic Branch Forking and Efficient Context Isolation.}
During inference, HPD-Parsing maintains a main layout branch that progressively predicts document regions in reading order. Whenever the layout branch emits a \texttt{<FORK>} token, the inference scheduler creates a localized content branch for the corresponding region. The new branch reuses the available visual and structural-prefix KV cache, replaces \texttt{<FORK>} with \texttt{<CHILD>}, and begins generating the regional content. Meanwhile, the main layout branch continues to identify subsequent regions, allowing multiple content branches to decode concurrently.

This design avoids redundant prefix computation. Each content branch shares the visual context and the layout prefix available at its forking position, while maintaining only its own incremental content KV cache. Consequently, a newly created branch does not need to re-encode the document image or repeat the common prefill computation. More importantly, each content branch is isolated from the detailed textual histories of unrelated regions. Under conventional full-page autoregressive decoding, every newly generated token attends to the complete preceding output sequence, causing the active KV cache and per-step attention cost to grow continuously with document length. In HPD-Parsing, each branch attends only to the shared visual context, its associated structural prefix, and its own localized generation history. This combination of shared-prefix reuse and localized context isolation shortens the active attention horizon and enables efficient concurrent decoding.
\subsection{Integration with P-MTP}
\label{sec:p-mtp}

Layout-coordinated parallel decoding removes unnecessary sequential dependencies across document regions, but token generation within each active branch remains autoregressive. To further shorten the intra-branch decoding trajectories, HPD-Parsing integrates Progressive Multi-Token Prediction (P-MTP) into both the main layout branch and localized content branches.

P-MTP employs a lightweight residual MLP to progressively predict multiple future tokens from the decoder hidden states. At decoding position $t$ and look-ahead depth $k$, the module combines the preceding hidden representation with the embedding of the previously predicted token, and projects the resulting feature through the shared LM head to obtain the speculative distribution $\hat{p}_{t}^{k}$. To stabilize predictions at increasing look-ahead depths, P-MTP adopts progressive loss weighting that assigns larger weights to reliable speculative trajectories. 

Under hierarchical parallel decoding, standard next-token prediction and progressive look-ahead prediction are jointly optimized across all decoding branches:
\begin{equation}
\label{eq:joint_pmtp}
\mathcal{L}
=
\sum_{k=0}^{K}
\frac{
\sum_{t=0}^{N}
\mathcal{M}_{t+k+1}
\mathcal{W}_{t,k}
\ell\!\left(\hat{p}_{t}^{k},y_{t+k+1}\right)
}{
\sum_{t=0}^{N}\mathcal{M}_{t+k+1}
},
\end{equation}
where $k=0$ denotes the standard next-token objective, while $k\geq1$ denotes progressive multi-token prediction. We set $\mathcal{W}_{t,0}=1$ for standard next-token prediction. For $k\geq1$, $\mathcal{W}_{t,k}$ denotes the progressive loss weight assigned to the depth-$k$ prediction at position $t$, which combines a distance-dependent decay with path- and target-consistency signals to down-weight unreliable long-range predictions as detailed in our previous work~\cite{xiang2026p}. The mask $\mathcal{M}_{t+k+1}$ specifies whether the target position is valid within the corresponding decoding branch. All valid positions are supervised for full-page and layout parsing, whereas content branches follow the content-validity rule defined in Equation~\ref{eq:dec_mask}.

During inference, P-MTP allows each active branch to draft multiple future tokens, verify them in parallel, and accept several tokens within a single decoding step. With an average accepted length of \textbf{6.6} tokens per step, it substantially reduces the decoding iterations required within both layout and content branches. Collectively, layout-coordinated branch concurrency and P-MTP shorten the decoding path across branches and within each branch, respectively.
\label{sec:method}

\section{Model Training and Data Curation}
HPD-Parsing is built on the open-source InternVL-3.5 1B and further adapted through a staged training strategy supported by a scalable data curation pipeline. This section first describes how the model transitions from conventional full-page parsing to hierarchical parallel decoding, and then introduces the construction and refinement of the corresponding training data.

\subsection{Staged Adaptation to Hierarchical Parallel Decoding}
\label{sec:train}
The training strategy of HPD-Parsing is designed to progressively acquire general document parsing capability, adapt the model to hierarchical parallel decoding, and improve task-specific output quality. Conventional full-page sequence generation provides dense supervision over complete parsing outputs and is therefore suitable for learning broad recognition, structural understanding, and reading-order modeling. However, hierarchical parallel decoding introduces distinct decoding roles for global layout coordination and localized content generation, which require branch-specific supervision. Moreover, the explicit decomposition of document outputs makes component-level reward assignment more tractable. Based on these considerations, we adopt a three-stage training strategy.

\begin{itemize}
    \item \textbf{Stage 1: Document Parsing Capacity Initialization.} 
        In the first stage, HPD-Parsing is trained using the conventional full-page sequence generation format to acquire general document parsing capability. Each document image is paired with its complete parsing output, providing dense supervision for text recognition, layout understanding, reading-order modeling, and structured generation. This formulation allows the model to learn from complete page-level sequences before being exposed to the hierarchical decoding format. Both the backbone and the P-MTP module are optimized under the joint objective in Equation~\ref{eq:joint_pmtp}, where all valid positions in the full-page output are supervised. This stage also enables P-MTP to acquire progressive look-ahead prediction capability from large-scale sequential outputs.

    \item \textbf{Stage 2: Decoding Paradigm Shift and Hard-case Optimization.} 
        Starting from the full-page parser initialized in Stage~1, the second stage adapts the model to hierarchical parallel decoding through branch-specific supervision. Each page is reorganized into two types of training instances corresponding to the decoding roles introduced in Section~\ref{sec:block}: global layout parsing and localized content decoding. Layout instances supervise the complete structural sequence, allowing the main branch to learn document organization, region assignment, and reading order. Content instances apply the validity mask in Equation~\ref{eq:dec_mask}, such that only the localized transcription following \texttt{<CHILD>} contributes to the objective. P-MTP is optimized under the same masks, extending each branch from next-token prediction to progressive look-ahead prediction. Since the model has already acquired general parsing capability, this stage primarily focuses on learning the new decoding format and strengthening performance on challenging document structures.

   \item \textbf{Stage 3: Reward-Guided Optimization.} 
       In the final stage, we perform lightweight reinforcement learning to improve output consistency and failure-prone parsing behaviors. The hierarchical output structure allows reward signals to be assigned to different parsing components more directly. We therefore construct task-aware rewards for formula quality, table parsing, and layout prediction, together with count-based consistency rewards. These signals provide fine-grained feedback on recognition fidelity, structural accuracy, and output conformity, enabling targeted optimization without substantially altering the general parsing capability acquired in the preceding stages.

\end{itemize}

\subsection{Automated Difficulty-Aware Data Curation Pipeline}
\label{sec:data}

\begin{figure}[t]
    \centering
    \includegraphics[width=1.02\linewidth]{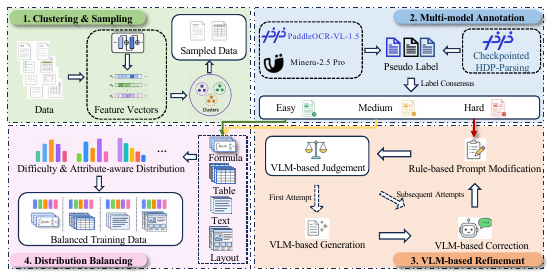}
    \caption{Overview of the automated difficulty-aware data curation pipeline, which consists of four stages: Clustering and Sampling, Multi-model Annotation and Difficulty Estimation, VLM-based Refinement and Distribution Balancing.}
    \label{fig:data}
\end{figure}

To construct diverse and reliable training data for the different training stages of HPD-Parsing, we develop an automated difficulty-aware data curation pipeline, as illustrated in Figure~\ref{fig:data}. The pipeline consists of four stages: feature-based clustering and sampling, multi-model annotation and difficulty estimation, VLM-based refinement, and distribution balancing.

\noindent\textbf{Clustering and Sampling.}
The first stage aims to improve data coverage while reducing redundancy. We extract visual feature representations from raw document images and group samples with similar document characteristics into clusters. Representative samples are then selected from different clusters, preventing the curated dataset from being dominated by highly repetitive layouts or visually similar pages. This process provides a diverse and scalable data pool for subsequent annotation.

\noindent\textbf{Multi-model Annotation and Difficulty Estimation.}
The sampled documents are annotated using complementary open-source document parsing models, including PaddleOCR-VL-1.5 and MinerU-2.5 Pro. Their predictions are used to construct initial pseudo-labels. Meanwhile, an intermediate HPD-Parsing checkpoint generates parsing results for the same samples. We evaluate the checkpoint predictions against the pseudo-labels using the OmniDocBench evaluation protocol and categorize samples into Easy, Medium, and Hard groups according to their parsing discrepancies. Samples that can be reliably parsed by the intermediate model are assigned to the easier groups, whereas samples exhibiting textual or structural errors are treated as hard cases.

\noindent\textbf{VLM-based Refinement.}
Hard samples are further processed through an iterative refinement procedure. A stronger VLM first examines the current annotation and identifies potential errors in text recognition, structure, or formatting. Based on the detected error types, the prompt is adjusted using predefined rules and the annotation is regenerated or corrected. The refined result is then evaluated again by the VLM. This process continues until the annotation satisfies the quality criterion or reaches a maximum of $N$ refinement attempts. The refinement stage is particularly beneficial for complex samples containing dense tables, multi-column layouts, formulas, or ambiguous region structures.

\noindent\textbf{Distribution Balancing.}
After annotation and refinement, the curated samples are balanced across both difficulty levels and document attributes. We regulate the proportions of Easy, Medium, and Hard samples to prevent the training data from being dominated by straightforward cases. We further balance the occurrence of major document elements and layout characteristics, including text blocks, tables, formulas, and complex layouts. The resulting dataset provides broader structural coverage and more reliable supervision for training under the hierarchical parallel decoding paradigm.


The pipeline is instantiated differently across the three training stages described in Section~\ref{sec:train}. For Stage~1, we construct 2.8 million representative full-page training samples through large-scale feature clustering, filtering, and sampling, with annotations generated primarily by MinerU-2.5 Pro. This stage emphasizes broad document coverage and scalable capability initialization. For Stage~2, we construct 100K branch-specific samples from the more thoroughly curated portion of the data pool. These samples undergo multi-model annotation, difficulty estimation, distribution balancing, and, for challenging cases, iterative refinement with a stronger VLM before being converted into the global layout parsing and localized content decoding formats. For Stage~3, we further select 600 representative hard cases from the refined data pool and organize them according to component-specific error patterns, providing targeted samples for task-aware reward optimization.

\label{sec:training}

\section{Inference Workflow}

To realize Hierarchical Parallel Decoding during inference, we extend the standard first-come-first-served (FCFS) scheduling framework with Layout-Coordinated Parallel Decoding and P-MTP. The resulting workflow dynamically forks concurrent content branches with shared-prefix KV cache reuse and accelerates token generation across all decoding branches, as formalized in Algorithm~\ref{alg:full_engine_final}.


\begin{algorithm2e}[H]
\SetKwProg{Fn}{Procedure}{}{}
\SetKw{KwAnd}{\textbf{and}}
\SetKw{KwBreak}{\textbf{break}}
\caption{FCFS Scheduling + Hierarchical Parallel Decoding}
\label{alg:full_engine_final}

\KwIn{KV occupancy threshold $\tau$, draft window $K$, concurrency cap $N_{\max}$ 
}
\KwOut{Decoded output sequences for all requests}
\KwData{
  $\mathcal{P}$: Active processing pool; \ 
  $\mathcal{Q}$: Waiting queue; \ 
  $\mathcal{M}[r]$: KV blocks of request $r$; \\
  $\textit{queue\_dirty} \in \{\mathtt{true}, \mathtt{false}\}$: Queue status flag; \ 
  $\operatorname{fork\_rel}[p]$: Active children of parent $p$
}
\While{\normalfont engine is running}{
  \tcp*[h]{FCFS Scheduling.} \\
  \If{\normalfont queue\_dirty}{
    \ForEach{$r \in \mathcal{Q}$}{
      \eIf{$r.\textit{is\_child}$}{
        $p \leftarrow r.\textit{parent\_id}$, \quad $m \leftarrow |\operatorname{fork\_rel}[p]|$\;
        $\textit{key}(r) \leftarrow (0,\; \textit{arrival}(p),\; -m,\; \textit{arrival}(r))$\;
      }{
        $\textit{key}(r) \leftarrow (1,\; \textit{arrival}(r))$\;
      }
    }
    $\mathcal{Q} \leftarrow \operatorname{sort\_by\_key}(\mathcal{Q})$;\quad $\textit{queue\_dirty} \leftarrow \mathtt{false}$\ \tcp*[r]{Sort waiting queue}
  }
  
  \ForEach{$r \in \mathcal{P}$}{
    \While{$\operatorname{AllocKV}(\mathcal{M}[r]) = \mathtt{null}$}{
      $v \leftarrow \arg\max_{r'\in\mathcal{P}}\, \textit{arrival}(r')$ \tcp*[r]{Preempt youngest}
      $\operatorname{PartialFree}(\mathcal{M}[v])$;\quad $\mathcal{Q}.\operatorname{prepend}(v)$\;
      \lIf{$v = r$}{\KwBreak}
    }
  }

  \While{$\mathcal{Q} \neq \emptyset$}{
    $r \leftarrow \mathcal{Q}.\operatorname{peek}()$\;
    \lIf{$\lnot r.\textit{is\_child}$ \KwAnd $(\operatorname{KVUsage}(\operatorname{KVPool}) > \tau \ \mathbf{or}\ |\mathcal{P}| \ge N_{\max})$}{\KwBreak}

    $\mathcal{Q}.\operatorname{pop}()$;\quad $\mathcal{P}.\operatorname{append}(r)$\tcp*[r]{Gate parents only; children bypass}
  }

  \tcp*[h]{P-MTP based Speculative Decoding \& Verification} \\
  \textbf{parallel for} $r \in \mathcal{P}$ \textbf{do} \ $r.\text{spec\_tokens} \leftarrow f_d(r.\text{context})$\;
  
  \ForEach{$r \in \mathcal{P}$}{
    $n_{\mathrm{acc}} \leftarrow \max\{k \mid \boldsymbol{y}^*_{n+i} = \hat{t}_i,\;\forall i \leq k\}$;\quad $r.\textit{len} \leftarrow r.\textit{len} + n_{\mathrm{acc}} + 1 - (K - n_{\mathrm{acc}})$\;
    $\operatorname{WriteKV}(\mathcal{M}[r],\; r.\textit{len})$\;
  }

  \vspace{1mm}
  \tcp*[h]{Fork Detection \& KV Zero-Copy Sharing} \\
  \ForEach{$r \in \mathcal{P}, \, t \in \boldsymbol{Y}_r$}{
    \If{$t = \textsc{ForkId} \ \mathbf{and}\ \lnot r.\textit{is\_child}$}{
      $c \leftarrow r.\operatorname{SpawnChild}()$;\quad $\operatorname{KVManager.fork\_ptr}(\mathcal{M}[r], \mathcal{M}[c])$\;
      $\mathcal{Q}.\operatorname{add}(c)$;\quad $\operatorname{fork\_rel}[r.\textit{id}].\operatorname{add}(c)$;\quad $\textit{queue\_dirty} \leftarrow \mathtt{true}$\;
    }
  }

  \vspace{1mm}
  \tcp*[h]{Reference-Counted Resource Release} \\
  \ForEach{$r \in \mathcal{P}$ \normalfont \textbf{with} $r.\textit{status} = \textsc{Done}$}{
    $\mathcal{P}.\operatorname{remove}(r)$\;
    \ForEach{$b \in \mathcal{M}[r]$}{
      $b.\textit{ref\_count} \leftarrow b.\textit{ref\_count} - 1$\;
      \lIf{$b.\textit{ref\_count} = 0$}{$\operatorname{KVPool}.\operatorname{free}(b)$}
    }
    \If{$r.\textit{is\_child}$}{
      $\operatorname{fork\_rel}[r.\textit{parent\_id}].\operatorname{remove}(r)$\;
    }
    \If{$|\operatorname{fork\_rel}[r.\textit{parent\_id}]| = 0$}{
      $\operatorname{Emit}(\operatorname{InterleaveOutputs}(r.\textit{parent\_id}))$\;
      $\operatorname{CleanupForkData}(r.\textit{parent\_id})$\;
    }
  }
}
\end{algorithm2e}

\section{Experiments}
\label{sec:exp}

\subsection{Experimental Setup}

\noindent\textbf{Implementation Details.}
We conduct all experiments on 8 NVIDIA A800 GPUs, each with 80GB of memory. Our model is built upon OpenGVLab/InternVL3.5-1B, and all parameters are optimized without freezing the vision encoder or aligner. The training process consists of three stages. In the first stage, Document Parsing Capacity Initialization, the model acquires general document parsing capabilities with an initial learning rate of $1 \times 10^{-4}$. In the second stage, Decoding Paradigm Shift and Hard-case Optimization, the model is adapted to the proposed hierarchical parallel decoding paradigm and further optimized on challenging cases using an initial learning rate of $1 \times 10^{-5}$. In the third stage,  Reward-Guided  Optimization., we further refine the model through reinforcement learning with a learning rate of $5 \times 10^{-7}$ and a global batch size of 96. We use bfloat16 precision throughout all three stages. For the first two stages, we train for 1 epoch, set the per-device batch size to 2, and use 8 gradient accumulation steps, resulting in an effective global batch size of 128. The maximum sequence length is set to 16,000 tokens, with a warmup ratio of 0.05. We adopt DeepSpeed ZeRO-1 with gradient checkpointing to reduce memory consumption and use FlashAttention for efficient attention computation.

For inference, we build our deployment system upon vLLM 0.17.1 and evaluate the throughout on NVIDIA A800 GPUs with 80GB memory. We implement the proposed hierarchical decoding mechanism within the vLLM serving framework, while preserving its first-come-first-served scheduling policy to maintain stable online serving behavior. The maximum generation length is set to 8,000 tokens for each decoding branch.

\noindent\textbf{Evaluation Metrics.}
We evaluate our model on widely-adopted document parsing benchmarks: OmniDocBench, which comprehensively assess document parsing capabilities across multiple dimensions including text recognition, formula recognition, table structure extraction, and reading order prediction. Following standard practices in document parsing literature, we adopt task-specific metrics for comprehensive evaluation: (1) \textbf{Text Edit Distance} (Edit$\downarrow$) measures character-level accuracy for text recognition; (2) \textbf{Formula CDM} (CDM$\uparrow$) evaluates mathematical formula recognition quality; (3) \textbf{Table TEDS} (TEDS$\uparrow$) and \textbf{Table TEDS-S} (TEDS-S$\uparrow$) assess table structure extraction accuracy with and without content recognition; (4) \textbf{Reading Order Edit Distance} (Edit$\downarrow$) quantifies the correctness of predicted reading sequences. The \textbf{Overall} score is computed as the weighted average across text, formula and table recognition.

\subsection{Main Results}

\begin{table*}[t]
\centering
\caption{Performance comparison on OmniDocBench v1.6. Best results in each category are in \textbf{bold}.}
\label{tab:effectiveness}
\resizebox{\textwidth}{!}{
\begin{tabular}{l|l|c|c|c|c|c|c|c}
\hline
\textbf{Model Type} & \textbf{Methods} & \textbf{Size} & \textbf{Overall$\uparrow$} & \textbf{TextEdit$\downarrow$} & \textbf{FormulaCDM$\uparrow$} & \textbf{TableTEDS$\uparrow$} & \textbf{TableTEDS-S$\uparrow$} & \textbf{ReadOrderEdit$\downarrow$} \\
\hline
\multirow{7}{*}{General VLMs} 
& InternVL3.5-241B~\cite{wang2025internvl3} & 241B & 83.76 & 0.130 & 89.95 & 74.35 & 79.78 & 0.215 \\
& Kimi K2.5~\cite{team2026kimi} & 1T & 84.53 & 0.107 & 83.50 & 80.76 & 84.00 & 0.211 \\
& GPT-5.2~\cite{singh2025openai} & - & 86.59 & 0.114 & 88.21 & 82.95 & 87.93 & 0.193 \\
& Qwen3-VL-235B~\cite{bai2025qwen3} & 235B & 89.78 & 0.063 & 92.55 & 83.07 & 86.75 & 0.166 \\
& Gemini 3 Flash~\cite{Gemini_3_2026} & - & 92.62 & 0.066 & 95.16 & 89.29 & 93.51 & 0.172 \\
& Gemini 3 Pro~\cite{Gemini_3_2026} & - & 92.91 & 0.064 & 95.99 & 89.15 & 92.96 & 0.165 \\
& Ovis2.6-30B-A3B~\cite{Ovis2_5_2025} & 30B & 93.70 & 0.035 & 95.17 & 89.44 & 92.40 & 0.135 \\
\hline
\multirow{11}{*}{\makecell[c]{Specialized VLMs \\(Pipeline)}} 
& Dolphin-1.5~\cite{feng2025dolphin} & 0.3B & 86.52 & 0.094 & 87.49 & 81.43 & 84.82 & 0.167 \\
& MonkeyOCR-pro-3B~\cite{MonkeyOCR_2025} & 3B & 88.57 & 0.074 & 88.74 & 84.35 & 88.62 & 0.189 \\
& Dolphin-v2~\cite{feng2026dolphin} & 3B & 89.50 & 0.069 & 91.01 & 84.40 & 87.44 & 0.150 \\
& OpenDoc-0.1B~\cite{du2025unirec} & 0.1B & 90.67 & 0.049 & 93.02 & 83.88 & 87.45 & 0.140 \\

& MinerU-2.5~\cite{MinerU_2_5_2025} & 1.2B & 93.04 & 0.045 & 95.77 & 87.88 & 91.47 & 0.130 \\
& Youtu-Parsing~\cite{yin2026youtu} & 2.5B & 93.74 & 0.044 & 93.63 & 92.02 & 95.00 & 0.116 \\
& PaddleOCR-VL~\cite{cui2025paddleocr} & 0.9B & 94.18 & 0.040 & 95.91 & 90.65 & 93.74 & 0.135 \\
& PaddleOCR-VL-1.5~\cite{cui2026paddleocr} & 0.9B & 94.93 & 0.038 & 96.89 & 91.67 & 94.37 & 0.130 \\
& GLM-OCR~\cite{duan2026glm} & 0.9B & 95.22 & 0.044 & 97.18 & 92.83 & 95.39 & 0.133 \\
& MinerU2.5-Pro~\cite{wang2026mineru2} & 1.2B & 95.75 & 0.036 & 97.45 & 93.42 & 95.92 & 0.120 \\

& PaddleOCR-VL-1.6~\cite{zhang2026paddleocr} & 0.9B & 96.3 & 0.032 & 97.5 & 94.8 & 97.11 & 0.127 \\

\hline
\multirow{11}{*}{\makecell[c]{Specialized VLMs \\(Unified)}} 
& POINTS-Reader~\cite{POINTS_reader_2025} & 3B & 83.37 & 0.096 & 85.72 & 73.98 & 77.40 & 0.198 \\
& Nanonets-OCR-s~\cite{NanonetsOCRs_2025} & 3B & 83.61 & 0.108 & 81.46 & 80.18 & 84.51 & 0.213 \\
& olmOCR~\cite{poznanski2025olmocr} & 7B & 85.74 & 0.139 & 88.10 & 83.00 & 87.17 & 0.216 \\
& OCRVerse~\cite{OCRVerse_2026} & 4B & 88.60 & 0.063 & 89.61 & 82.44 & 86.27 & 0.163 \\
& DeepSeek-OCR 2~\cite{wei2026deepseek} & 3B-A0.5B & 90.25 & 0.050 & 91.84 & 83.89 & 87.75 & 0.144 \\
& dots.ocr~\cite{li2025dots} & 3B & 90.77 & 0.048 & 89.95 & 87.18 & 90.58 & 0.138 \\
& FireRed-OCR~\cite{wu2026firered} & 2B & 93.26 & \textbf{0.037} & 95.44 & 88.04 & 91.06 & 0.131 \\

& Logics-Parsing-v2~\cite{chen2025logics} & 4B & 93.33 & 0.041 & 95.65 & 88.42 & 91.98 & 0.137 \\

& Qianfan-OCR~\cite{dong2026qianfan} & 4B & 93.90 & 0.040 & 95.08 & 90.53 & 93.31 & 0.130 \\

& Unlimited OCR~\cite{yin2026unlimited} & 3B-A0.5B & 93.92 & 0.042 & \underline{95.79} & 90.16 & 93.32 & 0.129 \\
& HunyuanOCR-1.5~\cite{li2026hunyuanocr} & 1B & \underline{94.74} & 0.039 & 94.50 & \textbf{93.67} & \textbf{94.71} & \underline{0.129} \\
& \textbf{Ours (HPD-Parsing)} & {1B} & \textbf{94.91} & \underline{0.039} & \textbf{97.28} & \underline{91.35} & \underline{94.11} & \textbf{0.124} \\
\hline
\end{tabular}
}
\end{table*}

\noindent\textbf{Effectiveness Comparison.}
Table~\ref{tab:effectiveness} compares HPD-Parsing with state-of-the-art document parsing methods on OmniDocBench v1.6. By combining global layout coordination with localized parallel decoding, HPD-Parsing preserves page-level structure while reducing the impact of difficult regions on the overall output. With only 1B parameters, it achieves an overall score of 94.91, establishing a new state of the art among end-to-end unified parsers and outperforming larger models such as Qianfan-OCR, Logics-Parsing-v2, and FireRed-OCR. It also remains competitive with leading pipeline-based parsers despite using substantially fewer training samples, achieving a ReadOrderEdit score of 0.124. This suggests that decomposing layout and content supervision enables more targeted optimization, while the unified architecture may better preserve global document context by allowing each region to be interpreted with awareness of the entire page.


\noindent\textbf{Efficiency Comparison.}
Table~\ref{tab:efficiency} compares the inference efficiency of different document parsing methods on OmniDocBench v1.6, together with their average input token counts. Compared to the autoregressive baseline, HPD-Parsing increases throughput from 1.02 to 2.68 PPS and from 1,554.8 to 4,752.1 TPS, corresponding to improvements $2.62\times$ and $3.06\times$, respectively. This substantial gain is achieved by replacing the conventional single autoregressive trajectory with hierarchical parallel decoding, which enables concurrent decoding across layout-coordinated content branches and further reduces the decoding steps within each branch through P-MTP. Input token count provides additional context for interpreting the efficiency comparison, as longer inputs generally incur greater prefilling and attention overhead. Despite processing approximately 4,800 input tokens per page, over four times that of DeepSeek-OCR-2, HPD-Parsing still achieves $1.31\times$ higher PPS and $1.62\times$ higher TPS, demonstrating strong inference efficiency under a considerably larger input-token budget.
\begin{table}[t]
\centering
\caption{Inference speed comparison on OmniDocBench v1.6. TPS(tokens per second), PPS(pages per second) under batch size 512.}


\label{tab:efficiency}
\resizebox{\columnwidth}{!}{
\begin{tabular}{l|l|c|c|c|c}
\hline
\textbf{Model Type} & \textbf{Method} & \textbf{Size} & \textbf{Avg Input Tokens} &  \textbf{TPS$\uparrow$ (BS=512)} & \textbf{PPS$\uparrow$ (BS=512)} \\
\hline
\multirow{4}{*}{\makecell[c]{Specialized VLMs \\(Pipeline)}} 
& Youtu-Parsing~\cite{yin2026youtu} & 2.5B & - & 315.4 & 0.39 \\
& MinerU2.5-Pro~\cite{wang2026mineru2} & 1.2B & - &1890.3 &  1.58 \\ 
& PaddleOCR-VL-1.6~\cite{zhang2026paddleocr} & 0.9B & - & 1533.8 & 1.25 \\
& GLM-OCR~\cite{duan2026glm} & 0.9B & - & 2133.8  & 1.86 \\ 
\hline
\multirow{7}{*}{\makecell[c]{Specialized VLMs \\(Unified)}} 
& Qianfan-OCR~\cite{dong2026qianfan} & 4B & 1856.9 &  994.47 & 0.60 \\
& FireRed-OCR~\cite{wu2026firered} & 2B & 4532.3 & 1082.0 & 0.90 \\
& HunyuanOCR-1.5~\cite{li2026hunyuanocr} & 1B & 4496.9 & 1081.3 & 1.10 \\
& DeepSeek-OCR-2~\cite{wei2026deepseek} & 3B-A0.5B & 1100.2 & 2932.1 & 2.05 \\ 
& Unlimited OCR~\cite{yin2026unlimited}  & 3B-A0.5B & 1485.6 & 2901.5 & 2.03\\ 
\cline{2-6}
& Baseline (Autoregressive) & 1B & 4809.3 & 1554.8 & 1.02 \\
& \textbf{Ours (HPD-Parsing}) & 1B & 4809.3 & \textbf{4752.1} & \textbf{2.68} \\
\hline
\end{tabular}
}
\end{table}

\begin{figure}[t]
\centering
\includegraphics[width=1\columnwidth]{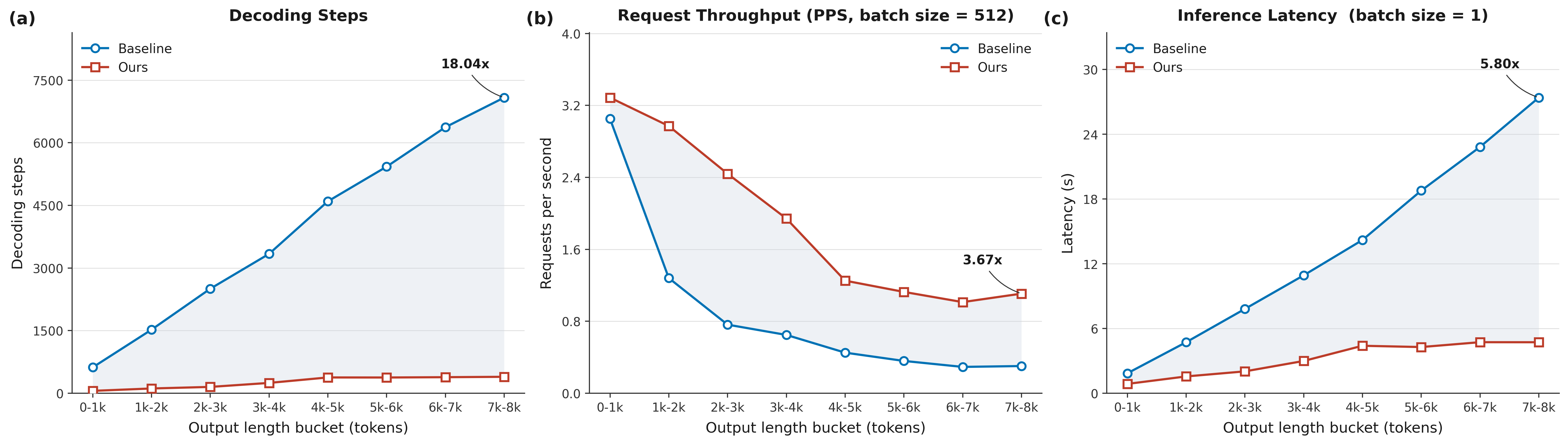}
\caption{\textbf{Efficiency scaling across output-length buckets.} Compared with the autoregressive baseline,  HPD-Parsing exhibits increasingly larger efficiency gains as output length grows. It reduces decoding steps by up to 18.04$\times$ in (a), improves request throughput by up to 3.67$\times$ under batch size 512 in (b), and reduces single-request inference latency by up to 5.80$\times$ under batch size 1 in (c).}
\label{fig:speed_vs_length}
\end{figure}

To examine how the acceleration gains vary with generation length, we partition the OmniDocBench v1.6 test set into output-length buckets and compare HPD-Parsing with the vanilla autoregressive baseline. As shown in Figure~\ref{fig:speed_vs_length}, the efficiency advantage becomes increasingly pronounced as the output sequence grows. Conventional unified document parsers generate the entire page along a single autoregressive trajectory, causing the number of decoding steps to increase approximately with the total output length. In contrast, HPD-Parsing uses the main layout branch for global coordination and delegates localized content decoding to concurrent branches. Consequently, the critical decoding path is primarily determined by the longest active branch rather than the sum of all block lengths. P-MTP further shortens each local decoding trajectory by generating and accepting multiple tokens at each step. By combining layout-coordinated branch concurrency with progressive multi-token prediction, HPD-Parsing alleviates both page-level sequential execution and token-by-token decoding within individual branches. Its acceleration advantage therefore grows with document length, reaching up to $18.04\times$ fewer decoding steps, $3.67\times$ higher request throughput, and $5.80\times$ lower single-request latency in the longest output-length bucket. These results demonstrate that hierarchical parallel decoding effectively mitigates the sequence-length-dependent slowdown of conventional autoregressive document parsing.

\begin{figure*}[t]
\centering
\vspace{-0.4cm}
\includegraphics[width=1\columnwidth]{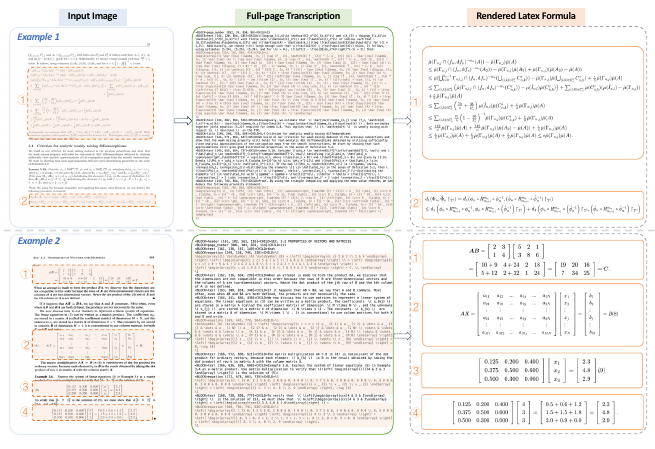}
\vspace{-0.8cm}
\caption{Qualitative results on document images with complex formulas.}
\vspace{-0.cm}
\label{fig:formula_samples}
\end{figure*}

\begin{figure}[!h]
\centering
\vspace{-0.2cm}
\includegraphics[width=1\textwidth]{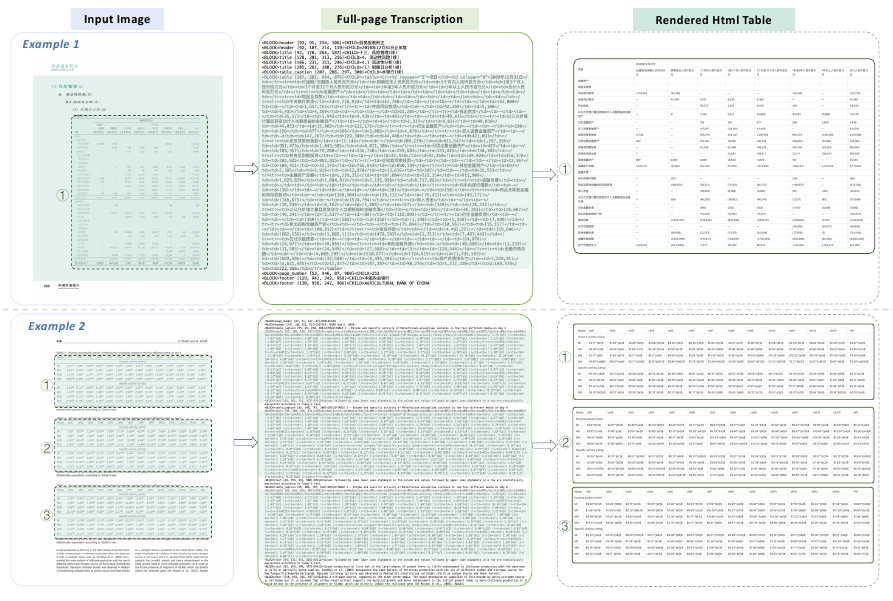}
\vspace{-0.5cm}
\caption{Qualitative results on document images with complex tables.}
\label{fig:table_samples}
\vspace{-0.2cm}
\end{figure}

\subsection{Qualitative Analysis}

\noindent\textbf{Comprehensive Coverage of Complex Documents.} 
Our method demonstrates robust parsing performance across document pages with diverse structural characteristics. Figure~\ref{fig:formula_samples} presents representative results on documents with complex mathematical expressions. The model accurately transcribes long multi-line formulas, matrix equations, fractions, and densely arranged mathematical symbols, while preserving valid LaTeX structures for faithful rendering. Figure~\ref{fig:table_samples} further evaluates the model on table-intensive documents. The results show that it can recover large and densely populated tables, including multi-row and multi-column structures, in a structured HTML representation while maintaining the original cell organization. Finally, Figure~\ref{fig:layout_samples} illustrates performance on pages with challenging layouts including newspapers and puzzle-style documents. The model preserves the spatial arrangement and reading order of interleaved text blocks, figures, and other visual elements, enabling coherent reconstruction of the overall page structure. Together, these qualitative examples show that the unified parsing framework can consistently process heterogeneous document content within full-page transcription.

\begin{figure*}[!h]
\centering
\vspace{-1cm}
\includegraphics[width=0.92\textwidth]{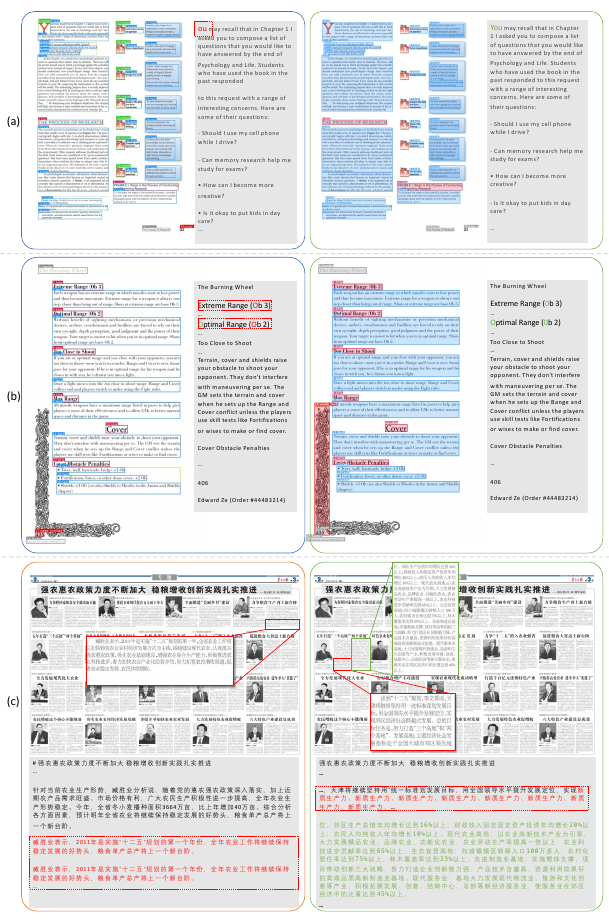}
\vspace{-0.3cm}
\caption{Comparative advantages over competing methods.  HPD-Parsing demonstrates superior robustness in (a) handling detection errors caused by misaligned bounding boxes, (b) recognizing visually similar characters through global contextual reasoning, and (c) preventing repetition errors from propagating across document blocks.}
\vspace{-0.3cm}
\label{fig:compare}
\end{figure*}

\noindent\textbf{Comparative Advantages over Competing Methods.}
We provide qualitative comparisons to illustrate the key advantages of our hierarchical parallel decoding paradigm over both traditional pipeline-based parsers and competing unified VLM approaches. Figure~\ref{fig:compare} presents representative failure cases of competing methods alongside our successful parsing results, highlighting three critical benefits:

\begin{figure}[!h]
\centering
\includegraphics[width=0.95\textwidth]{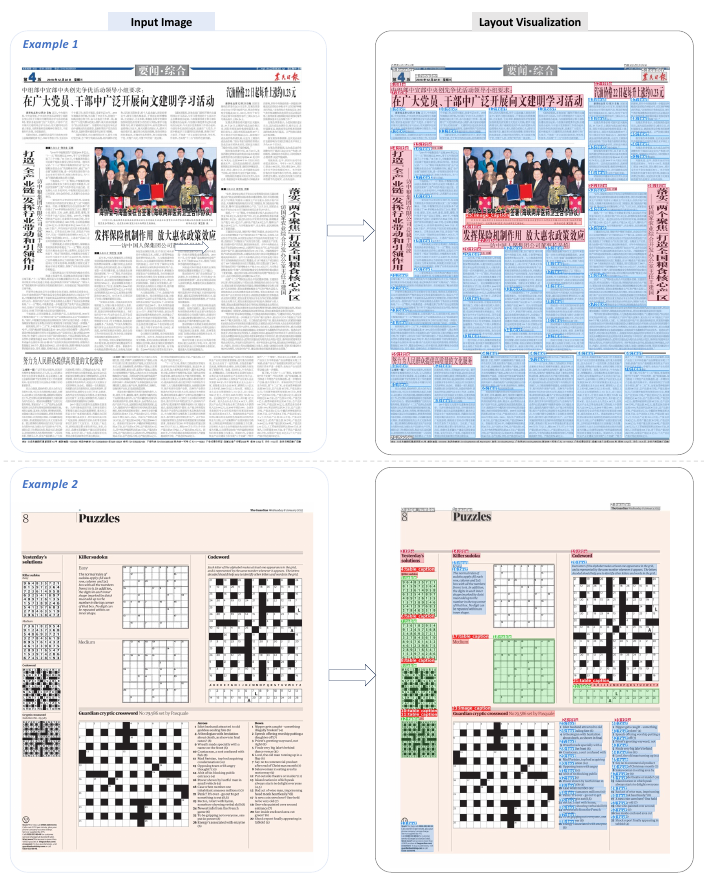}
\caption{Qualitative results on document images with complex layout.}
\label{fig:layout_samples}
\end{figure}

\begin{itemize}
\item \textbf{Robustness to Detection Errors.}
Pipeline-based methods depend heavily on the layout analysis stage. As shown in Figure~\ref{fig:compare}(a), inaccurate or incomplete bounding boxes may exclude relevant content or mix adjacent regions, directly causing recognition errors in the subsequent stage. In contrast,  HPD-Parsing incorporates document-level visual and semantic context during parsing, enabling correct recognition despite imperfect region localization.

\item\textbf{Handling Visually Similar Characters.}
Figure~\ref{fig:compare}(b) presents a representative case involving the visually similar characters `0' and `O'. Pipeline-based methods process isolated text regions with limited contextual information and may therefore confuse the digit zero with the uppercase letter O. By leveraging broader document context,  HPD-Parsing more reliably disambiguates the two characters and produces the correct parsing result.

\item\textbf{Error Propagation Prevention.}
Unified autoregressive methods may enter repetitive decoding once an error occurs, causing the corrupted output to propagate through subsequent document content. As illustrated in Figure~\ref{fig:compare}(c), competing methods repeatedly generate the same text span. In  HPD-Parsing, localized content is decoded in independent branches, confining repetition errors to individual regions and preventing them from affecting subsequent document blocks.
\end{itemize}

\section{Conclusion and Future Work}


In this paper, we present HPD-Parsing, a lightweight and high-throughput document parser built upon a hierarchical parallel decoding paradigm. Motivated by the observation that document parsing requires global coordination while content decoding is largely localized, HPD-Parsing replaces the traditional single autoregressive trajectory with a main layout branch for global coordination and concurrent local decoding branches for region-level content generation. Progressive Multi-Token Prediction further reduces the decoding steps within both the layout and content branches. Experiments on public benchmarks demonstrate that HPD-Parsing substantially improves inference throughput while maintaining competitive parsing accuracy, validating layout-coordinated parallel concurrency as an effective alternative to fully sequential document generation. More generally, the globally coordinated yet locally decomposable structure of documents indicates that hierarchical parallel decoding may extend beyond single-page parsing to key information extraction, multi-page document understanding, and other structured generation tasks. In the future, we will further expand the training data to improve coverage across a wider range of document scenarios and explore structure-aware attention designs that reduce the effective attention context and the computation required at each decoding step.
\clearpage

\newpage


{\small
\bibliography{main}
\label{sec:ref}
}

\newpage
\setlength{\cftbeforesecskip}{6pt}   
\setlength{\cftbeforesubsecskip}{4pt} 
\setcounter{tocdepth}{2}

\newpage
\end{document}